\documentclass{article}

\usepackage{neurips_2024}

\usepackage[utf8]{inputenc} 
\usepackage[T1]{fontenc}    
\usepackage{hyperref}       
\usepackage{url}            
\usepackage{booktabs}       
\usepackage{amsfonts}       
\usepackage{nicefrac}       
\usepackage{microtype}      
\usepackage{xcolor}         
\usepackage{amsmath,bm}
\usepackage{float}
\usepackage{multirow, graphicx}
\usepackage{subfigure}
\usepackage{graphicx}
\usepackage{tabularx, makecell, multirow} 
\usepackage{wrapfig}
\usepackage{amsmath,amssymb}

\title{NGM-SLAM: Gaussian Splatting SLAM with Radiance Field Submap}

\author{%
  David S.~Hippocampus\thanks{Use footnote for providing further information
    about author (webpage, alternative address)---\emph{not} for acknowledging
    funding agencies.} \\
  Department of Computer Science\\
  Cranberry-Lemon University\\
  Pittsburgh, PA 15213 \\
  \texttt{hippo@cs.cranberry-lemon.edu} \\
}

\begin{document}
\bibliographystyle{plain}

\maketitle

\begin{abstract}

Gaussian Splatting has garnered widespread attention due to its exceptional performance. Consequently, SLAM systems based on Gaussian Splatting have emerged, leveraging its capabilities for rapid real-time rendering and high-fidelity mapping. However, current Gaussian Splatting SLAM systems usually struggle with large scene representation and lack effective loop closure adjustments and scene generalization capabilities. To address these issues, we introduce NGM-SLAM, the first GS-SLAM system that utilizes neural radiance field submaps for progressive scene expression, effectively integrating the strengths of neural radiance fields and 3D Gaussian Splatting. We have developed neural implicit submaps as supervision and achieve high-quality scene expression and online loop closure adjustments through Gaussian rendering of fused submaps. Our results on multiple real-world scenes and large-scale scene datasets demonstrate that our method can achieve accurate gap filling and high-quality scene expression, supporting both monocular, stereo, and RGB-D inputs, and achieving state-of-the-art scene reconstruction and tracking performance.

\end{abstract}

\section{Introduction}
SLAM systems\cite{pire2017s, newcombe2011dtam, mur2015orb, liao2022so, Qin2018, qin2018vinsmono} have long been a fundamental concern in the domains of robotics and AR/VR. Dense SLAM\cite{schops2019bad, huang2021di, runz2017co, bloesch2018codeslam, craig2004tandem, teed2021droid}, in particular, commands a broader range of applications and demand compared to its sparse counterparts. Traditional dense SLAM systems\cite{whelan2012kintinuous, deng2023long} utilize explicit representations like voxels, point clouds, and TSDF\cite{yao2019recurrent}, and have achieved commendable results in tracking. However, the limitations of traditional dense SLAM\cite{ Whelan2015, Hosseinzadeh2019} in high-fidelity modeling, gap filling, and texture details restrict their wider application.

Neural implicit SLAM, leveraging NERF-based\cite{yu2021pixelnerf, deng2022depthsupervisednerf, guo2022nerfren, barron2021mip} implicit representations, has crafted a complete pipeline that includes tracking and mapping, significantly enhancing the perceptual capabilities of SLAM systems\cite{deng2023long}. Despite these advancements, current neural implicit SLAM systems face limitations in operational speed, real-time capabilities, and memory requirements. Recent developments in SLAM methods based on 3D Gaussian Splatting (3DGS)\cite{kerbl2023gaussian} not only retain the high-fidelity mapping benefits of NERF-SLAM systems but also demonstrate advantages in rendering speed and precision, challenging the dominant position of neural implicit SLAM in dense reconstruction. 3DGS employs continuous Gaussian volumes characterized by color opacity and orientation for scene representation, combining the intuitive flexibility of explicit representations with the continuous differentiability of implicit expressions.

However, it's notable that 3DGS, compared to NERF\cite{rosinol2022nerf, muller2022instant, wang2021neus, turki2022mega}, shows limitations in gap-filling capabilities mainly due to its reliance on point cloud models as the primary input and lacks the generalizing inference capabilities from neural networks. Additionally, current 3DGS-SLAM systems lack robust loop detection, which is crucial for correcting global drift errors and preventing map loss. This disadvantage becomes particularly apparent in large scenes due to cumulative drift.

To address these challenges, we propose a SLAM system based on neural submaps and 3DGS representation. We first establish submaps represented by neural radiance fields and construct a global keyframe list. When triggering map thresholds, we create new neural submaps and use previous submaps as priors to guide Gaussian rendering. Subsequently, we perform local bundle adjustment (BA). Between submaps, we implement a submap fusion strategy and trim the generated submaps' Gaussians. Upon loop closure detection, we perform real-time coarse-to-fine global loop adjustments, adjusting the map poses corresponding to anchor frames and applying global BA and global BA Gaussian rendering loss to correct poses and associated local map keyframes. We significantly mitigate errors caused by accumulated drift and map drift, while correcting global map loops with minimal computational overhead. Experimental results demonstrate that our method achieves state-of-the-art performance in tracking and mapping and is scalable to large-scale scenes. In summary, our contributions are as follows:

\begin{itemize}
  \item We introduce the first progressive dense Gaussian splatting SLAM system based on neural submaps, achieving high-fidelity mapping through a local-to-global reconstruction strategy. We enable large-scale scene inference and effectively leverage the advantages of both representation methods.
  \item We propose a global loop correction strategy, including coarse-to-fine submap correction and global bundle adjustment loss, enabling real-time adjustment of submaps and map correction.
  \item We propose effective Gaussian pruning and multiscale Gaussian rendering strategies, ensuring the system can remove redundant Gaussians while enhancing anti-aliasing capabilities, improving rendering speed, and accuracy.
  \item Our system supports monocular, stereo and RGB-D inputs, demonstrating competitive tracking and mapping performance on 5 datasets, and supports real-time inference at 5 FPS in large-scale scenes.
\end{itemize}

\section{Related Work}

\textbf{Dense Visual SLAM} Dense real-time scene mapping is considered a key method for solving problems related to scene understanding, autonomous driving, and AR/VR applications\cite{deng2023prosgnerf}. Early methods like KinectFusion utilized an explicit scene representation approach to achieve comprehensive tracking and mapping, offering more accurate scene expression and geometric shapes compared to sparse point-based methods. Traditional dense SLAM systems extensively employ various explicit representations such as voxels and point clouds. The advantages of traditional dense SLAM methods lie in their accurate and mature tracking systems, but they have limitations in providing high-fidelity models and lack generalized reasoning capabilities.

Recent years have seen significant attention given to methods based on neural radiance fields\cite{zakharov2020autolabeling,wang2022nerf,ichnowski2021dex,hu2022nerf}. iMAP\cite{sucar2021imap} established the first complete neural implicit SLAM framework, achieving scalable and efficient scene representation. However, limitations due to a single MLP architecture can lead to tracking loss and mapping errors in larger-scale scenes. NICE-SLAM\cite{zhu2022nice} utilizes a pretrained multi-MLP system with frozen parameters to achieve accurate tracking and scene expression, yet it faces drift issues in filling gaps. ESLAM\cite{bloesch2018codeslam} uses tri-plane features for scene representation, while Go-SLAM\cite{zhang2023goslam} supports multimodal inputs from monocular, stereo, and RGB-D cameras. However, these neural implicit methods\cite{zhu2023nicerslam} generally lack online loop closure correction, often resulting in the loss of high-frequency details due to local over-smoothing.

\textbf{3D Gaussian Based SLAM} Recently, 3D Gaussian based scene representation methods have garnered broad interest. Compared to NERF-based neural implicit methods\cite{kong2023vmap, Li2023, chung2023orbeezslam}, 3D Gaussian methods\cite{yan2023gs} combine the advantages of explicit and implicit expressions. 
 They capture high-fidelity three-dimensional scenes through a differentiable rasterization process, avoiding the per-pixel ray casting required by neural fields, thus achieving high-speed rendering. 
Photo-SLAM\cite{hhuang2024photoslam} achieves high-quality real-time representation using a Gaussian pyramid-based approach. SplaTAM\cite{keetha2024splatam} employs anisotropic Gaussian representation, achieving real-time tracking and rendering. MonoGS{Matsuki:Murai:etal:CVPR2024} achieves faster scene representation through Gaussian shape regularization and geometric verification, but lacks loop closure correction to eliminate accumulated errors. However, these methods face many challenges. Because 3D Gaussian (3DGS) representations cannot learn scene features for inference, they cannot fill gaps like NERF-based methods. Furthermore, current 3DGS SLAM methods still lack stable and complete tracking systems, such as loop detection and online bundle adjustment (BA). We propose an incremental scene representation method that combines the strengths of NeRF and 3DGS for complementary scene expression—a method that can learn features while retaining complete high-frequency details. By establishing a neural implicit submap based on NERF as a prior to guide Gaussian rendering, our method can effectively fill gaps and achieve more comprehensive scene reconstruction. The use of multi-scale Gaussian rendering ensures optimized rendering speed and enhances local details. The online adjustment of our neural submap allows global BA without the need for re-rendering. Although our method adopts a hybrid rendering approach, our parallel submap tracking and optimization avoid the costs associated with global scene updates, thereby improving overall system speed.

\section{Method}

We present the system pipeline of NGM-SLAM in Figure 1. Our system first tracks and constructs submaps based on RGB/RGB-D image streams using neural radiation fields (Sec 3.1). Then, the neural submaps are utilized to construct Gaussian priors supervised submaps (Section 3.2). Following alignment of submap poses, we execute a submap fusion strategy (Sec 3.3). Finally, we design a local-to-global loop closure detection and bundle optimization process (Sec 3.4). We will elaborate on the entire process of our system in the methodology.

\begin{figure*}[h]
\centering
\includegraphics[width=0.9\textwidth,height=0.55\textwidth]{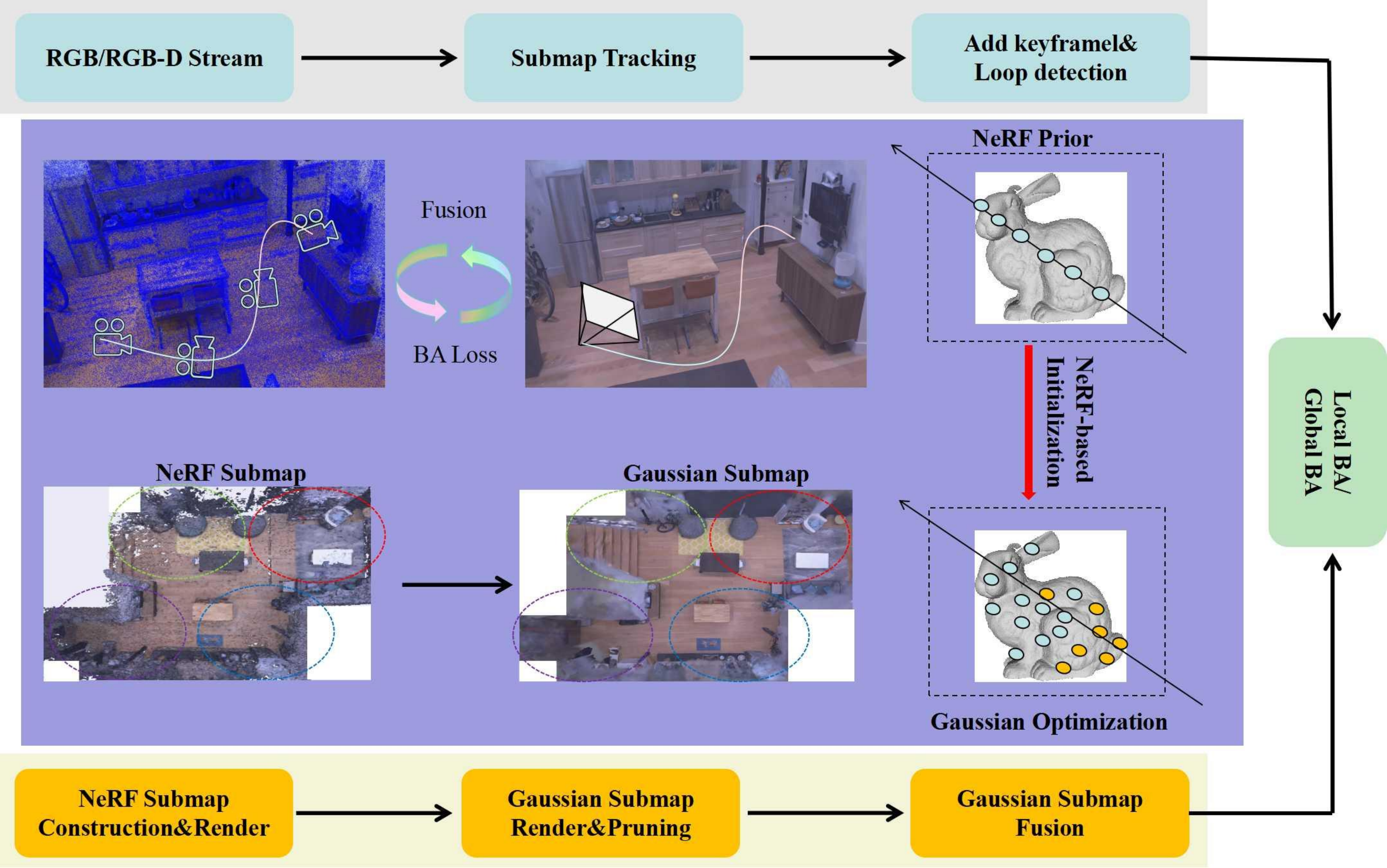}
\caption{{The system includes two modules: tracking and mapping. After the initial submap starts to be established, the tracking module continuously estimates the camera pose and detects loops, while passing keyframes of the submap to the mapping module. The mapping module first constructs a neural submap that also serves as a prior for the multi-scale GS(Gaussian Splatting) submap, and performs parallel rendering between submaps. Local Bundle Adjustment (BA) is conducted within submaps to correct pose and mapping errors, and Global BA is executed on all anchor frames when a loop closure is detected. Finally, the resulting GS maps are stitched together.}}
\label{fig2}
\end{figure*}

\subsection*{3.1 Neural Submap Construction}
\textbf{Neural Submap Construction} In current 3D Gaussian-based SLAM systems, due to the lack of generalized inferencing capabilities of neural networks, we have introduced data-driven, incremental neural submaps as a foundational supervisory mechanism to fill voids and enhance map representation. Our approach is based on ORB feature point tracking\cite{campos2021orb}, as feature point extraction is performed only on the current input frame, which avoids the frequent drift issues associated with submap creation and hole filling.

Initially, we establish a local submap list of keyframes and a global list of keyframes. This facilitates the effective implementation of local/global Bundle Adjustment (BA) processes. When tracking begins, the first frame of the local submap is simultaneously added to the local keyframes and global keyframes list. We set a keyframe threshold for submaps; when a submap accumulates enough keyframes to reach this threshold, we establish a new local submap.

To better achieve map fusion and avoid excessive fusion errors between submaps, we utilize the DBOW model to assess the co-visibility relationships between keyframes. If the map connection frames lack co-visibility, the current submap's connecting frame is added as the first frame to a newly created submap to ensure stable map integration.

\textbf{Neural Implicit Rendering} Due to the significantly higher speed of Gaussian rendering compared to neural implicit rendering, we adopt a progressive rendering strategy to achieve real-time rendering between submaps and the global map. We initially perform neural implicit rendering using only a sparse set of keyframes for each submap to obtain foundational supervision. Once the submap is preliminarily adjusted, full-frame poses are utilized for Gaussian rendering to refine the mapping process. To reduce the rendering cost, we employ a multi-resolution hash-encoded radiance field as a prior.

The radiance field $f$ is a continuous function that maps a three-dimensional point position $p \in \mathbb{R}^3$ and a viewing direction $d \in S^2$ to a volume density $\sigma \in \mathbb{R}_+$ and a NeRF RGB color value $\mathbf{C} \in \mathbb{R}^3$. Inspired by NeRF volume rendering, the final color prediction for a pixel is approximated through ray marching and integration using sample points:
\begin{equation}
\mathbf{C}_{\mathrm{N}} = \sum_{j=1}^{N} T_{Nj} \alpha_j \mathbf{C}_j, \quad \text{where} \quad T_{Nj} = \prod_{k=1}^{j-1}(1-\alpha_k), \quad \alpha_j = 1-e^{-\sigma_j \delta_j}
\end{equation}
where $T_{Nj}$ is the NeRF transmittance, $\alpha_j$ is the alpha value for $x_j$, and $\delta_j$ is the distance between adjacent sample points. In neural radiance fields, $f$ is parameterized as an MLP with ReLU activation $f_\theta$, and the network parameters $\theta$ are optimized via gradient descent on the reconstruction loss:
\begin{equation}
\mathcal{L}(\theta) = \sum_{\mathbf{r} \in \mathcal{R}_{\text{b}}}\left\|\mathbf{C}_{\mathrm{N}}^\theta(\mathbf{r}) - \mathbf{C}_{\mathrm{GT}}(\mathbf{r})\right\|^2
\end{equation}
where $r \in \mathcal{R}_{\text{b}}$ is a batch of rays sampled from the set of all rays.

\subsection*{3.2 Gaussian Submap}

\textbf{Multi-scale Gaussian Rendering} Utilizing neural submap priors, we represent the scene using a set of anisotropic 3D Gaussian distributions.The scene is depicted as points associated with a position \( p \in \mathbb{R}^3 \), opacity \( o \in [0,1] \), third-order spherical harmonics (SH) coefficients \( k \in \mathbb{R}^{16} \), a 3D scale \( s \in \mathbb{R}^3 \), and a 3D rotation \( R \in \text{SO}(3) \) represented by a 4D quaternion \( q \in \mathbb{R}^3 \). ${\mathbf{c}}$ is the color value when rendering in 3D Gaussian. Inspired by equation (1), this representation can be rendered to a camera's image plane, with a correctly ordered list of points:
\begin{equation}
\mathbf{c}_{\mathrm{GS}} = \sum_{j=1}^{N_p} \mathbf{c}_j \alpha_j T_{Gi} \quad \text{where} \quad T_{Gi} = \prod_{i=1}^{j-1}\left(1-\alpha_i\right)
\end{equation}
Where $T_{Gj}$ is the 3D Gaussian transmittance. Each Gaussian \( G_i \) contains optical properties: color \( \mathbf{c}_i \) and opacity \( \alpha_i \). For a continuous 3D representation, the mean \( \overline{x}_i \) and covariance \( \textbf{C}_i  \) defined in world coordinates describe the Gaussian's location and its ellipsoidal shape. The 3D Gaussians \( N(\overline{x}_W, \textbf{Z}_W) \) in world coordinates are associated with 2D Gaussians \(N(\overline{x}_I, \textbf{Z}_I) \) on the image plane through the projection transformation:
\begin{equation}
\boldsymbol{\overline{x}}_I = \pi\left(\boldsymbol{T}_{CW} \cdot \boldsymbol{\overline{x}}_W\right), \quad \mathbf{\textbf{Z}}_I = \mathbf{J}W \mathbf{\textbf{Z}}_W \mathbf{W}^T \mathbf{J}^T
\end{equation}
$W$ is the viewing transformation, $\mathbf{J}$ denotes the Jacobian of the affine approximation of the projective transformation \cite{zwicker2001ewa}, and ${\mathbf{Z}}_W$ denotes the 3D covariance matrix. The radiance fields provided by neural submaps serve as the foundational supervision for rendering, but they are prone to aliasing effects that degrade the rendering quality during the sampling process. This is particularly evident when constructing submaps with many small Gaussians, leading to severe artifacts. We employ a multi-scale Gaussian rendering approach inspired by \cite{Yu2024GOF}, aggregating smaller Gaussians into larger ones to improve the rendering quality. We represent from Gaussian functions at four detail levels corresponding to 1×, 4×, 16×, and 64× down-sampling resolutions. During the training process, smaller fine-level Gaussian functions are aggregated to create coarser-level larger Gaussians. The selection of Gaussian bodies is based on pixel coverage, including or excluding according to the coverage range defined by the inverse of the highest frequency component in that region \( f_{\text{max}} = \frac{1}{S_k} \). Specifically, Gaussians at the edges of submaps are not aggregated to facilitate submap fusion operations and submap alignment.

\textbf{Ray-Guided Gaussian Pruning} To reduce the number of ineffective Gaussian volumes during Gaussian rendering, thereby improving rendering speed, we adopted a pruning method in the Gaussian rendering process guided by ray sampling. We employed an importance assessment strategy to remove invalid points from all the Gaussians. The importance score is defined by the contribution of the aggregated Gaussian particles to the rays in all input images. To improve filtering efficiency, we introduced a sparse point cloud composed of ORB map points as guidance, which typically corresponds to regions near textured surfaces. Inspired by~\cite{niemeyer2024radsplat}, we used a point cloud counter to gather statistics and counted rays with a number of nearby sparse points exceeding the threshold $t_1$ as the ray set $ K_r $. The importance score can be expressed as:
\begin{equation}
\textbf{E}(\mathbf{p}_i) = \max_{I_f \in \mathcal{I}_f, r \in I_f,r \in \textbf{K}_r} (\alpha_i^r T_i^r )
\end{equation}
where $I_f$ is the rendered image and $\mathcal{I}_f$ is the target image. \( \alpha_i^r \tau_i^r \) represents the contribution of Gaussian \( i \) to the final color prediction of a pixel, as described in equation (4). The mask values are computed as follows: 
\begin{equation}
m_i = m(\mathbf{p}_i) = \mathbf{1}(\textbf{E}(\mathbf{p}_i) < t_{\text{prune}})
\end{equation}
where \( t_{\text{prune}} \in [0,1] \) is a threshold used to control the number of points representing the scene. All Gaussian distributions with a mask value of 1 are removed from the scene. Finally, we execute default rasterization, ensuring that our rendering speed does not decrease and improving accuracy through pruning.

\subsection*{3.3 Submap Fusion}

We represent the scene as the sum of multiple local scenes: 
\begin{equation}
\left\{I_i, D_i\right\}_{i=1}^M \mapsto\left\{\mathrm{SF}_{\sigma_1}^1, \mathrm{SF}_{\sigma_2}^2, \ldots, \mathrm{SF}_{\sigma_n}^n\right\}
\end{equation}
The series $\left\{I_i, D_i\right\}_{i=1}^M$ represents a sequence of RGB-D inputs, where $SF_{\sigma_n}^n$ denotes the submap representation.
When generating a new submap, all submaps are anchored based on the spatial positions of local keyframe poses. After each local Bundle Adjustment (BA), the central pose of the map is adjusted for re-anchoring. To avoid overlapping artifacts at the edges of rendered submaps, we remove Gaussian bodies outside all submap boundaries, effectively reducing boundary artifacts. Then, we proceed with submap stitching. To ensure seamless maps, we apply the Gaussian aggregation method described in Sec 3.2, aggregating smaller Gaussians at map boundaries into larger ones. We observe that the merged boundaries are seamless. After loop closure adjustment and global BA execution, we repeat the map fusion process. Our submap fusion strategy avoiding excessive memory consumption due to continuous map expansion.

\subsection*{3.4 Loop Closure and BA}
\textbf{Local-Global Loop Closure} To correct accumulative drift, we perform local Bundle Adjustment (BA) within each submap, involving only local submap keyframe corrections. Inspired by \cite{campos2021orb}, we utilize the Bag of Words (BoW) model for relevance detection among global keyframes. When loop closure conditions are met, a global optimization process is initiated. To align global submaps and fuse submaps, we adopt a coarse-to-fine global adjustment strategy.

 In contrast to traditional methods, we first optimize the pose of anchor frames using BA and perform a submap fusion process. This prevents drift at the boundaries of Gaussian submaps. Afterwards, we fix the position of anchor frames, execute a global BA process based on the global keyframe list, and then perform a second Gaussian submap fusion process to complete loop closure. This enables our system to correct significant drift while avoiding missing and overlapping artifacts caused by map misalignment. Additionally, we randomly sample rays from all keyframes involved in global BA to guide the generation and fusion of Gaussian bodies, implementing the process described in Section 3.2 to further correct rendering errors.

\textbf{Local-Global Bundle Adjustment} Unlike methods that adjust using neural point clouds, our approach does not correct all mapping errors at once. Instead, we perform coarse-to-fine rendering adjustments. Thanks to the speed advantage of 3D Gaussian rendering, we can achieve real-time re-rendering and construct bundle adjustment (BA) losses.

To ensure geometric shape and appearance consistency, we distort the rendered RGB and depth to the co-visible keyframes, constructing the loss according to the following equations:
\begin{align}
\mathcal{L}_{\mathrm{BA}-\text{Igb}} & = \sum_{i=1}^{N-1} \sum_{j=i+1}^N \left(\left|T_i^j \cdot \mathcal{R}(T_i, c) - F_j^c\right|\right) \\
\mathcal{L}_{\mathrm{BA}-\text{depth}} & = \sum_{i=1}^{N-1} \sum_{j=i+1}^N \left(\left|T_i^j \cdot \mathcal{R}(T_i, d) - F_j^d\right|\right)
\end{align}
where \( F_j^c \) and \( F_j^d \) are the color and depth of keyframe \( j \), respectively. \( \mathcal{R}(T_i, c) \) and \( \mathcal{R}(T_i, d) \) represent the rendered RGB and depth. Thus, the overall loss function \( \mathcal{L}_{\mathrm{BA}} \) used for joint optimization of the corresponding keyframe poses and 3D Gaussian scene representation is a weighted sum of the above losses. In our experiments, we found that the scale explosion of aggregated Gaussian bodies during the BA process could affect rendering. Therefore, we apply an $\mathcal{L}_2$ loss \(\mathcal{L}_{rgs}\) to 3D Gaussians with scales exceeding the threshold $t_2$. The total rendering loss \(\mathcal{L}\) we obtain is:
\begin{equation}
\mathcal{L}_{\mathrm{BA}} = \lambda_1{L}_{\text{color}} + \lambda_2 \mathcal{L}_{\text{depth}} + \lambda_3 \mathcal{L}_{\text{rgs}} 
\end{equation}

Where \( \lambda_1 \), \( \lambda_2 \), and \( \lambda_3 \) are weighting coefficients. To address potential local errors caused by forgetting Gaussian submaps, we conduct extra optimization iterations for all co-visible keyframes. Unlike methods using neural point clouds, our approach benefits from faster rendering and lower computational costs of Gaussian rasterization. Unlike the high computational cost in \cite{zhang2023goslam}, we avoid globally re-rendering neural radiation fields.

\section{Experiments}

\textbf{Implementation Details.} We implemented NGM-SLAM on a desktop computer equipped with an Intel i7-12700K and an NVIDIA RTX 3090 Ti with 24 GB. Our implementation utilized mixed programming in C++ and Python. To ensure fair comparisons, we provided experimental data for both monocular and RGB-D setups. Detailed parameter settings are available in the supplementary materials. Our baselines include traditional methods, neural implicit approaches, and state-of-the-art (SOTA) systems based on 3D Gaussian SLAM, including ORBSLAM3\cite{campos2021orb}, BAD-SLAM\cite{schops2019bad}, Vox-Fusion\cite{yang2022voxfusion}, DROID-SLAM\cite{teed2021droid}, Co-SLAM\cite{wang2023coslam}, ESLAM\cite{bloesch2018codeslam}, Go-SLAM\cite{zhang2023goslam}, Point-SLAM\cite{Sandstrom2023PointSLAM}, SplaTAM\cite{keetha2024splatam} and MonoGS\cite{Matsuki:Murai:etal:CVPR2024}. For some data, we used the results reported in the respective papers of these methodologies.

\textbf{Datasets and Metrics.} We utilized four RGB-D datasets, which include 8 small room sequences and 4 large-scale multi-room sequences from the Replica\cite{straub2019replica} dataset, featuring complex corridors and stairs. Additionally, we used 3 indoor scene sequences captured by three real sensors from the TUM RGB-D\cite{sturm2012evaluating} dataset, 6 sequences from the ScanNet\cite{dai2017scannet} dataset, comprising large-scale real indoor scenes, 4 indoor scene sequences from the ScanNet++\cite{Yeshwanth_2023_ICCV} dataset, and 3 indoor sequences with challenging large perspective changes from the EuRoC\cite{Burri2016} dataset. For tracking, we employed ATE RMSE (cm) as the benchmark. For reconstruction, we compared rendering accuracy using PSNR, SSIM, and LPIPS. Our computational results represent the average of five experiments rendered along the camera's direction for all frames. We assessed running speed and computational resource requirements using FPS and GPU Memory Usage.

\subsection {Evaluation on Replica}

\begin{table*}[h]
\begin{threeparttable}
\renewcommand\arraystretch{1.3} 
\resizebox{\textwidth}{!}{
\begin{tabular}{c|cccccccccccc}
\hline
\multirow{1}{*}{Metrics}                                                                                     & PSNR(dB)↑  &SSIM↑  &LPIPS↓ &ATE(cm)↓  &Tracking FPS↑ &System FPS↑ &GPU Usage(G)↓   \\ \hline
\multirow{1}{*}{NICE-SLAM \cite{zhu2022nice}}                                                        
           &24.42     &0.81   &0.23 &2.35 & 2.33 &1.91 &6.27   \\

\multirow{1}{*}{Co-SLAM \cite{wang2023coslam}}                                                         
            &30.24     &0.86  &0.18 &1.16 &14.58 & \textbf{12.64} &\textbf{5.83}    \\

\multirow{1}{*}{Go-SLAM \cite{wang2023coslam}}              
 &24.15    & 0.77  &0.35 &1.12 &10.74 & \underline{8.26} &14.44   \\ 

\multirow{1}{*}{Point-SLAM \cite{Sandstrom2023PointSLAM}}                                                            
 &33.49     &\underline{0.97}   &0.14 &0.73  & 1.10 &0.42  &7.31  \\

\multirow{1}{*}{SplaTAM \cite{keetha2024splatam}}                                                           
 &31.81     &0.96   &0.16 &\underline{0.55} &1.07 &0.42 &18.87   \\ 

\multirow{1}{*}{MonoGS \cite{Matsuki:Murai:etal:CVPR2024}}                                                               
 &34.05    &0.96   &\underline{0.12} &{0.58} &4.58 &2.26 &27.99    \\ 

\multirow{1}{*}{NGM-SLAM(Mono)}                                                          
 &\underline{35.02}     &{0.96}   &{0.13} &{8.51} &\underline{16.11} &3.82 &{7.62}\\ 

\multirow{1}{*}{NGM-SLAM}                                                          
 &\textbf{37.43}     &\textbf{0.98}   &\textbf{0.08} &\textbf{0.51} &\textbf{20.54} &5.71 &\underline{5.98}    \\  \hline
\end{tabular}}
\end{threeparttable} 
\caption {{The average results of five measurements for eight scenes of a sequence of smaller rooms in the Replica\cite{straub2019replica} dataset are reported for PSNR (dB), SSIM, LPIPS, ATE (cm), Tracking FPS, System FPS, and GPU usage. The best results are bolded, and the second best results are indicated with an underline.}}\label{tab1}      
\end{table*}

As shown in Table 1, we provide experimental data for monocular and RGB-D small-scale room scenes, comparing them with other methods. We demonstrate the quality evaluation of scene reconstruction, including PSNR(dB), SSIM, LPIPS, and ATE RMS (cm), as well as the running speed. Our method achieves state-of-the-art results, ensuring real-time performance while running on lower memory, which gives it an advantage over other 3DGS-SLAM methods. It serves as a foundation for extension to mobile platforms such as robots. Our monocular mode can also achieve reconstruction quality close to RGB-D without depth supervision. As shown in Figure 1, we compare the reconstruction results of monocular and RGB-D modes with other methods on room sequences from the Replica\cite{straub2019replica} dataset. Compared to the baseline method, our reconstruction shows better detail, clearer texture, fills holes, and avoids ghosting and local reconstruction errors.

\begin{figure*}[h]
\centering
\includegraphics[width=1.0\textwidth,height=0.55\textwidth]{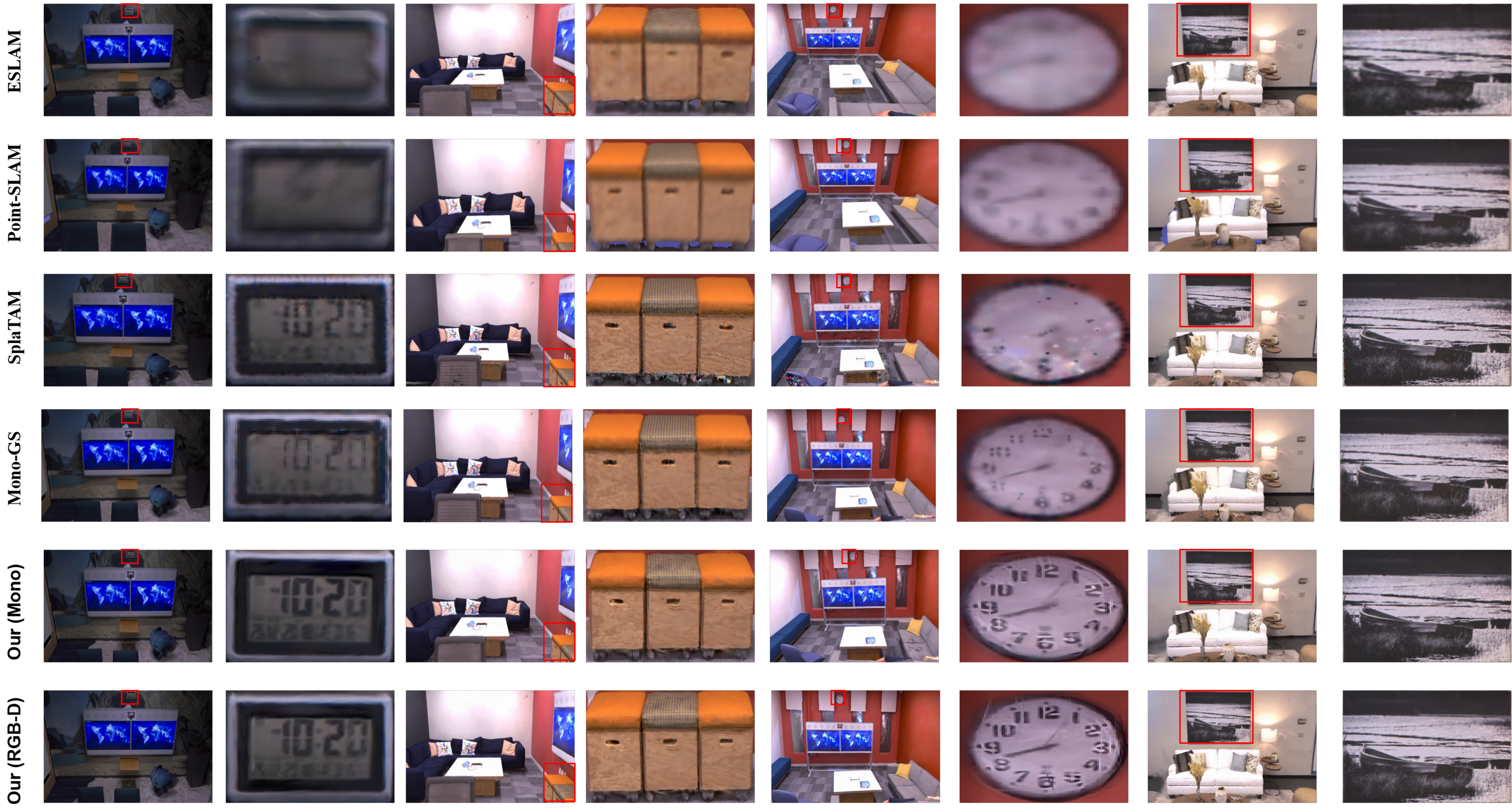}
\caption{{We present scene and local detail results on four sequences in the Replica\cite{straub2019replica} dataset, including monocular and RGB-D reconstruction. Our method exhibits superior detail expression and overall reconstruction, while preserving the finest texture details.}}
\label{fig2}
\end{figure*}

\begin{table*}[h]
\begin{threeparttable}
\renewcommand\arraystretch{1.3} 
\resizebox{\textwidth}{!}{
\begin{tabular}{c|cccccccccccc}
\hline
\multirow{1}{*}{Part}                               & Apartment-0         & Apartment-1  &Apartmen-2  &Frl-apartment-0 &Frl-apartment-4  &Average \\ \hline

\multirow{1}{*}{NICE-SLAM \cite{zhu2022nice}}                                                        
           &16.99     &14.52   &8.19 &2.01 &2.37 &8.46   \\

\multirow{1}{*}{Co-SLAM \cite{wang2023coslam}}                                                         
            &\underline{8.60}     &9.78  &6.31 &1.94 &0.81 &5.89   \\

\multirow{1}{*}{Go-SLAM \cite{zhang2023goslam}}              
 &14.10    &\textbf{3.54}  &\textbf{1.00} &\textbf{0.40} &\textbf{0.12} & \underline{3.23}  \\ 

\multirow{1}{*}{Go-SLAM(Mono) \cite{zhang2023goslam}}                                                           
 &29.81     &17.43   &5.39 &1.50 &2.02 &11.23  \\ 

\multirow{1}{*}{Point-SLAM \cite{Sandstrom2023PointSLAM}}                                                            
 &13.49     &10.97   &8.22 &2.21  & 1.48 & 7.27    \\ 

\multirow{1}{*}{MonoGS \cite{Matsuki:Murai:etal:CVPR2024}}                                                               
 &19.02    &9.37   &2.52 &0.98 &0.97 &6.57  \\ 

\multirow{1}{*}{NGM-SLAM(Mono)}                                                          
 &{17.86}     &{21.18}   &{13.89} &{8.61} &{5.66} &13.24    \\  

\multirow{1}{*}{NGM-SLAM}                                                          
 &\textbf{5.23}     &\underline{4.91}   &\underline{1.01} &\underline{0.81} &\underline{0.33} &\textbf{2.46}   \\  \hline
\end{tabular}}
\end{threeparttable} 
\caption {{The performance of ATE RMSE (cm) on 5 large-scale scene sequences from the Replica\cite{straub2019replica} dataset. The average of 5 measurements is taken for each sequence. The best result is indicated in bold, and the second-best result is underlined. Our method outperformed the baseline method.}}\label{tab1}      
\end{table*}

As shown in Table 2, we present the tracking results on five larger-scale scenes from the Replica\cite{straub2019replica} dataset. Our method's ATE results show a 23.9\% improvement in accuracy compared to Go-SLAM\cite{zhang2023goslam}. As illustrated in Figure 2, compared to the baseline method, we avoid the accumulation of errors caused by large-scale tracking. In complex large-scale scenes such as Apartment-1, which includes multiple rooms and corridors between rooms, accumulated errors can lead to catastrophic forgetting of the scene, highlighting the importance of loop closure. Meanwhile, we ensure advantages in detail representation and reasonable hole filling.

\begin{figure*}[h]
\centering
\includegraphics[width=1\textwidth,height=0.3\textwidth]{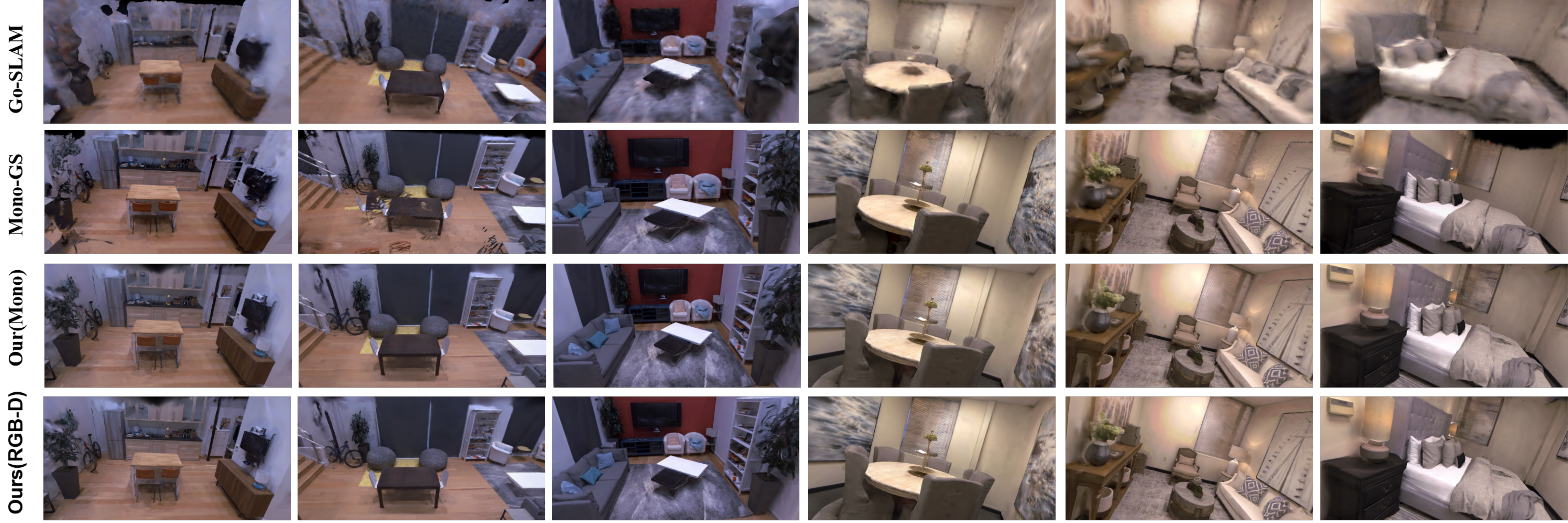}
\caption{{The reconstruction results on four large-scale apartment sequences, each consisting of multiple rooms, in the Replica\cite{straub2019replica} dataset demonstrate that our method achieves more accurate reconstruction compared to Nerf-based approaches and state-of-the-art MonoGS\cite{Matsuki:Murai:etal:CVPR2024}. It avoids catastrophic forgetting. Moreover, as demonstrated in the final sequence showcasing window details, we can achieve reasonable background completion and scene generalization.}}
\label{fig2}
\end{figure*}

\subsection {Evaluation on ScanNet}

\begin{table*}[h]
\begin{threeparttable}
\renewcommand\arraystretch{1.3} 
\resizebox{\textwidth}{!}{
\begin{tabular}{c|cccccccccccc}
\hline
\multirow{1}{*}{Part}            &scene0000           &scene0059  &scene0106 &scene0169 &scene0181   &scene0207 &Average \\ \hline

\multirow{1}{*}{NICE-SLAM \cite{zhu2022nice}}              
&8.64 &12.25 &8.09 &10.28 &12.93 &5.59 &9.63    \\ 

\multirow{1}{*}{Co-SLAM \cite{wang2023coslam}}                                                        
           &\underline{7.13}    &11.14   &9.36 &5.90 & 11.81 & \underline{7.14}  &8.75 \\

\multirow{1}{*}{Vox-Fusion \cite{yang2022voxfusion}}                                                         
            &16.62     &24.23  &8.41 &27.33 &23.31 &9.49  &18.23 \\

\multirow{1}{*}{ESLAM \cite{bloesch2018codeslam}}              
  &7.54   & 8.52  & {7.39} & 8.17& \textbf{9.13} & \textbf{5.61} & \underline{7.73}        \\  

\multirow{1}{*}{Point-SLAM \cite{Sandstrom2023PointSLAM}}                                                           
 &10.24     &7.81   &8.65 &22.16 &14.77 &9.54  &12.20\\ 

\multirow{1}{*}{SplaTAM \cite{keetha2024splatam}}                                                            
  &12.83   &10.10 &17.72  &12.08 &11.10 & 7.47   &11.88    \\

\multirow{1}{*}{NGM-SLAM(w/o loop)}                                                          
 &{7.42}     &\underline{6.43}   &\underline{7.31} &\underline{6.81} &{12.33} &{7.92}  &{8.05} \\  

\multirow{1}{*}{NGM-SLAM(w/ loop)}                                                          
  &\underline{6.71}  & \textbf{6.26}   &\textbf{7.24} & \textbf{5.83}  & \underline{10.12}    &7.44     &\textbf{7.27}    \\\hline
\end{tabular}}
\end{threeparttable} 
\caption {{The performance of ATE RMSE (cm) on 5 large-scale scene sequences from the Replica\cite{straub2019replica} dataset. The average of 5 measurements is taken for each sequence. The best result is indicated in bold, and the second-best result is underlined. Our method outperformed the baseline method.}}\label{tab1}      
\end{table*}

Our tracking results on the ScanNet\cite{dai2017scannet} dataset are shown in Table 3. We demonstrate more robust tracking when loop closure detection and bundle optimization are enabled. In Figure 2, we illustrate the local reconstruction results, where our method can reasonably fill in gaps, such as chairs and walls. Additionally, by eliminating the accumulation errors in mapping and correcting mapping errors caused by incorrect scene updates, we can accurately recover the geometric shapes of objects, such as bicycles, shoes, and cellos in the scene.

\begin{figure*}[h]
\centering
\includegraphics[width=0.9\textwidth,height=0.6\textwidth]{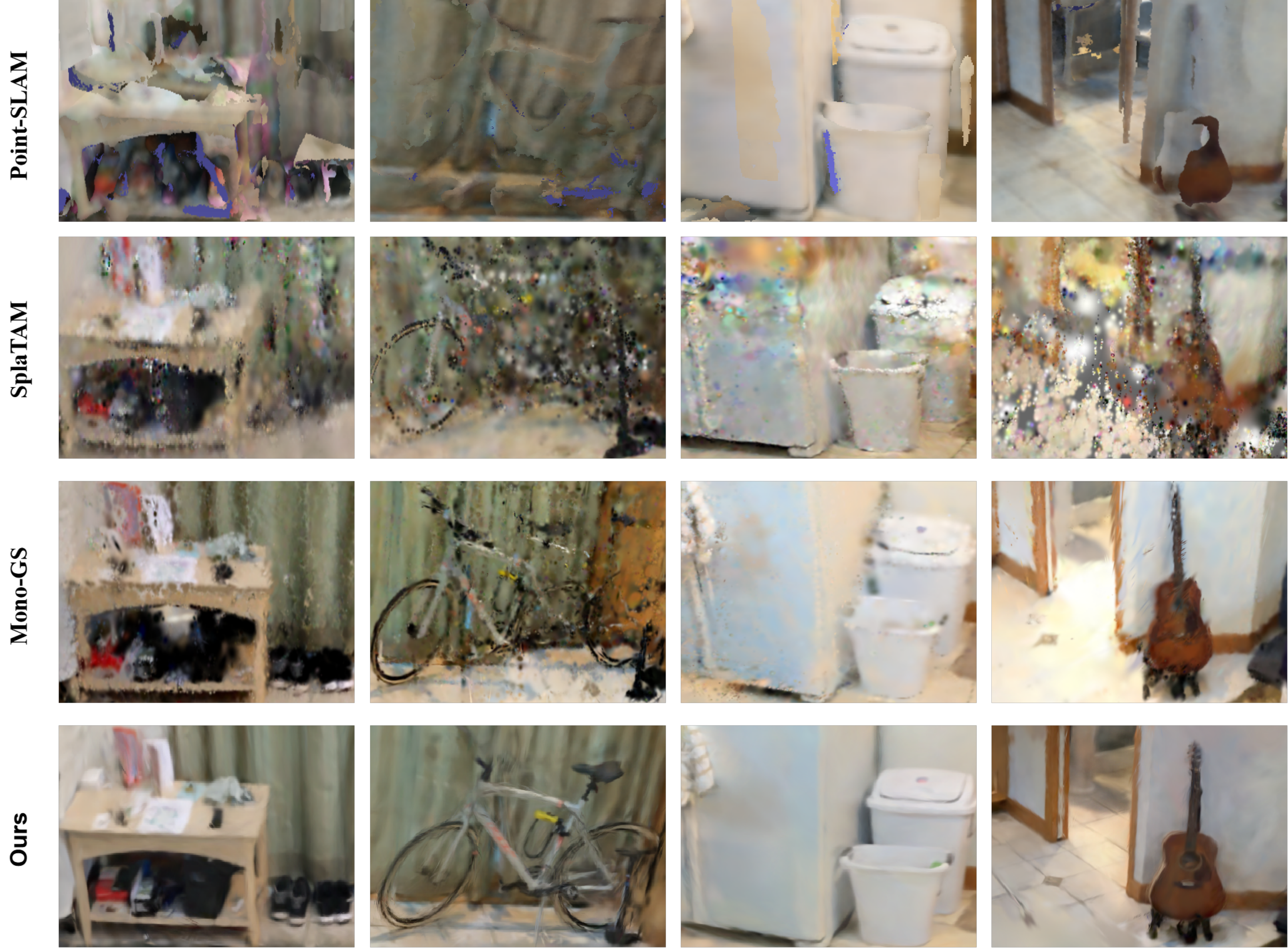}
\caption{{On large-scale multi-room sequences in the ScanNet dataset, our method demonstrates superior error accumulation correction capability compared to current 3DGS-based approaches. We can accurately ensure consistency across multiple views, avoiding erroneous scene reconstructions such as blurry shoes and bicycles, while also preventing local detail collapse caused by aliasing artifacts.}}
\label{fig2}
\end{figure*}

\subsection {Evaluation on TUM RGB-D and EuRoC}
Our ATE RMSE (cm) and construction results in indoor scenes for TUM RGB-D\cite{sturm2012evaluating} datasets as shown in Tables 4 and EuRoC\cite{Burri2016} dataset in Tables 5. Our method achieves competitive results compared to traditional approaches in tracking performance.

\begin{table*}[h]
\begin{minipage}[b]{0.48\textwidth}
\begin{adjustbox}{width=\textwidth}
\centering
\begin{threeparttable}
\renewcommand\arraystretch{1.3}
\begin{tabular}{c|cccc}
\hline
                              & fr1\_desk & fr2\_xyz & fr3\_office & AVG \\ \hline
DI-Fusion \cite{huang2021di}   & 4.4       & 2.0      & 5.8         & 4.1 \\
BAD-SLAM \cite{schops2019bad}  & 1.7       & 1.1      & 1.7         & 1.5 \\
ESLAM \cite{bloesch2018codeslam} & 2.3       & 1.1      & 2.4         & 2.0 \\
MonoGS \cite{Matsuki:Murai:etal:CVPR2024} & \textbf{1.52}  & 1.58     & 1.65        & 1.58 \\
NGM-SLAM(Mono)                 & \underline{1.57}  & \underline{0.72} & \underline{1.42} & \underline{1.24} \\
NGM-SLAM                       & 1.72      & \textbf{0.40} & \textbf{1.00} & \textbf{1.04} \\ \hline
\end{tabular}
\caption{The performance of ATE RMSE (cm) on TUM RGB-D\cite{sturm2012evaluating} dataset. The best result is indicated in bold, and the second-best result is underlined.}
\label{tab:left}
\end{threeparttable}
\end{adjustbox}
\end{minipage}%
\hfill
\begin{minipage}[b]{0.41\textwidth}
\begin{adjustbox}{width=\textwidth}
\centering
\begin{threeparttable}
\renewcommand\arraystretch{1.3}
\begin{tabular}{c|cccc}
\hline
                              & V101 & V102 & V103 & AVG \\ \hline
ORB-SLAM2 \cite{mur2015orb}   & 0.035  & 0.020 & 0.048 & 0.034 \\
ORB-SLAM3 \cite{campos2021orb} & 0.035  & 0.025 & 0.052 & 0.037 \\
SVO \cite{forster2016svo} & 0.045 & 0.040 & 0.070 & 0.051 \\
Go-SLAM \cite{zhang2023goslam} & {0.041} & 0.040 & 0.024 & 0.035 \\
DROID-SLAM  \cite{teed2021droid}   & \underline{0.037} & \textbf{0.026} & \underline{0.023} & \underline{0.029} \\
NGM-SLAM                       & \textbf{0.033}  & \underline{0.027} & \textbf{0.020} & \textbf{0.027} \\ \hline
\end{tabular}
\caption{The performance of ATE RMSE (cm) for stereo input on the EuRoC\cite{Burri2016} dataset, with the best result indicated in bold and the second-best result underlined.}
\label{tab:right}
\end{threeparttable}
\end{adjustbox}
\end{minipage}
\end{table*}

\section{Conclusion}
We propose NGM-SLAM, the first Gaussian SLAM system based on neural submaps. Through the priors provided by neural submaps and loop closure adjustments, we achieve high-quality tracking and reconstruction of large-scale scenes, enabling local-to-global loop closures and correcting cumulative errors. We strike a balance between high-quality texture detail representation and real-time operation. Experimental results demonstrate that our approach outperforms state-of-the-art NERF/GS-based SLAM methods in terms of rendering and tracking accuracy.

\end{document}